\documentclass{article}

\usepackage[breaklinks=true,colorlinks,bookmarks=false]{hyperref}
\usepackage[legalpaper, margin=1in]{geometry}
\usepackage[utf8]{inputenc} 
\usepackage[T1]{fontenc}    
\usepackage{times}
\usepackage{url}            
\usepackage{booktabs}       
\usepackage{amsfonts}       
\usepackage{nicefrac}       
\usepackage{microtype}      %
\usepackage{graphicx}
\usepackage{amsmath,amssymb}
\usepackage{subcaption}
\usepackage{wrapfig}
\usepackage{cite}
\usepackage{multirow}

\usepackage[linesnumbered,ruled,vlined]{algorithm2e}

\SetCommentSty{mycommfont}
\SetKwInput{KwInput}{Input}               
\SetKwInput{KwOutput}{Output}
\SetKwInput{KwInit}{Initializations}

\title{Self-Supervision by Prediction for Object Discovery in Videos}

\begin{document}
\author{Beril Besbinar, Pascal Frossard\\
LTS4 - EPFL\\
{\tt\small {name.surname}@epfl.ch}}
\date{} 
\maketitle

\begin{abstract}
Despite their irresistible success, deep learning algorithms still heavily rely on annotated data.
On the other hand, unsupervised settings pose many challenges, especially about determining the right inductive bias in diverse scenarios.
One scalable solution is to make the model generate the supervision for itself by leveraging some part of the input data, which is known as self-supervised learning.
In this paper, we use the prediction task as self-supervision and build a novel object-centric model for image sequence representation.
In addition to disentangling the notion of objects and the motion dynamics, our compositional structure explicitly handles occlusion and inpaints inferred objects and background for the composition of the predicted frame.
With the aid of auxiliary loss functions that promote spatially and temporally consistent object representations, our self-supervised framework can be trained without the help of any manual annotation or pretrained network.
Initial experiments confirm that the proposed pipeline is a promising step towards object-centric video prediction.

\end{abstract}

\section{Introduction}
\label{sec:intro}
Most of today's deep neural networks lack notions of compositionality and modularity as well as capability of reasoning.
The vast majority of existing methods in computer vision and robotics are trained in a supervised way to leverage at best the datasets without paying particular attention to the notion of objects despite its crucial importance.
Not only does this limit their use to the domain the algorithm is trained on, but it also leads to representations that are tied to a specific task.
In contrast, humans learn about objects by observing or interacting \cite{baillargeon2004infants,gopnik1999scientist} and reuse this information after gradual refinements.
This means that sequences of images with moving objects may provide an ideal playground for learning object representations, provided that one can model a physically plausible scene structure. 

In this work, we take a new approach towards self-supervised object discovery in videos, by proposing compositional and order-invariant representations based on temporal consistency. 
The proposed model is heavily inspired by \textit{common fate} of objects as well as \textit{spatial proximity} \cite{goldstein2009perceiving}.
We represent objects by masks, which are obtained by seeking for compact sets of pixels that obey the same simplified geometric transformation model throughout the sequence. 
We learn the permutation of masks that determines the proper relative depth ordering of objects and ensure occlusion-aware composition of a frame by generating a pixel occupancy map based on the ordered object masks.
Upon decomposition of the current frame, we use inpainting modules to update the appearance clues for each object, as well as a separate background model.
We then predict transformation parameters for each object and compose the predicted frame using the objects warped with the predicted set of parameters.
The proposed model is implemented in an end-to-end differentiable way to be trained with auxiliary loss terms, which support our video model assumptions and aid joint training of the different modules.
Our initial results demonstrate that our new model provides a promising direction towards the inference of effective compositional, object-centric representations, as well as the video prediction task.

Video prediction has been a well-studied topic in the last years \cite{ranzato2014video, mathieu2015deep, walker2016uncertain, lotter2016deep, finn2016unsupervised, kalchbrenner2017video, liu2017video, villegas2017decomposing, vondrick2017generating, liang2017dual, lu2017flexible, xu2018structure, hsieh2018learning, tulyakov2018mocogan, gao2019disentangling, villegas2019high, weissenborn2019scaling, kumar2019videoflow, ye2019compositional, wu2020future, zablotskaia2020unsupervised, fan2019cubic, jia2016dynamic, xue2016visual, walker2015dense, castrejon2019improved, franceschi2020stochastic, denton2017unsupervised, lee2018stochastic}.
However, recent works mostly focus on modelling the distribution of data in order to infer all plausible outcomes given initial frame(s), rather than studying the underlying factors of variation, such as the structure of the scene or how objects move and interact with each other. 
Some methods \cite{villegas2017decomposing, hsieh2018learning, tulyakov2018mocogan} decompose frames into 'content' and 'motion' to attribute stochasticity to the 'motion' component.
However, the 'content' part is rarely object-centric or does not consider invariant representations.
These methods mostly process a frame in a sequential manner where the order of processing is of crucial importance and the resultant object representations are not invariant to the imposed order or processing.
For the generation of predicted frame(s), the majority of recent models rely on pixel-wise transformations.
They are very often explicitly estimated as optical flow \cite{liang2017dual, lu2017flexible, gao2019disentangling} resulting in high fidelity outcome for real video sequences, or sometimes with Spatial transformers \cite{jaderberg2015spatial} as in \cite{finn2016unsupervised, vondrick2017generating}. 
Explicit optical flow estimation mostly implies the use of pre-trained models, and dependence on annotated datasets.
Spatial Transformers, on the other hand, enable joint learning of motion in prediction frameworks by lowering the dimension of search space with a compromise in accuracy, when used for local regions, or objects. 
Otherwise, when used for per-pixel transformations as in \cite{vondrick2017generating}, they lead to high capacity models accompanied by high computational resource requirement.
Closer to our proposal, Ye et al. \cite{ye2019compositional} follow a compositional approach by factorizing abstract visual entities, yet, they operate in the latent space rather than with visual clues and their method depends on the initial localization of entities.
Wu et al. \cite{wu2020future} proposes a very similar pipeline to ours; however, it builds on pre-trained networks for many sub-tasks although the method is claimed to be unsupervised.
In contrast, we propose to work in fully unsupervised settings.
Moreover, we attempt to explicitly infer the full support of the projection of each \textit{moving} object to the image plane, which is the first such proposal for video decomposition and prediction, to the best of our knowledge.
Although our method might seem very close to video object segmentation (VOS) at first glance, it differentiates from VOS mainly in its purpose but also in its methodology.
Indeed, even in unsupervised settings, methods for VOS rely on annotations for training, while our method does not require any kind of annotation and it is purely data-driven.
Our method rather participates to an increasing interest in very recent years for object-centric learning in images \cite{greff2017neural, van2018relational, greff2019multi, locatello2020object}, with most methods employing attention and iterative algorithms.
While our method can be considered as a different interpretation of attention, the non-iterative nature of our formulation results in lower computational cost at inference time.

In the following sections, we introduce our formulation for object-centric video prediction and present our implementation, as well as the  preliminary results. 
Section \ref{sec:formulation} describes the formulation with particular emphasis to the sub-modules for frame decomposition, temporal modelling by parametric spatial transformations, inpainting and composition of the predicted frame.
Section \ref{sec:implementation} explains how the functions described in Section \ref{sec:formulation} are implemented in an end-to-end differentiable fashion while introducing the auxiliary loss terms that assist the proposed pipeline to learn targeted representations.
Section \ref{sec:experiments} presents the datasets used for training and testing the proposed method, as well as the corresponding results, both qualitatively and quantitatively.
Section \ref{sec:conclusion} concludes the paper by summarizing the crucial outcomes and discussing future work.

\section{Video Representation}
\label{sec:formulation}

In this section, we first introduce our video model assumptions.
Later, we present the formulation of our compositional model.
We then describe our implementation in the following section, with particular attention to the loss function that permits our self-supervised scheme.

\subsection{Model Assumptions}
Our new video representation model is based on some simplifying yet still generic assumptions that:
\begin{itemize}
    \setlength{\itemsep}{0pt}
    \setlength{\parskip}{0pt}
    \item[(i)] a video frame is composed of objects whose motion between consecutive frames can be approximated by simple geometric transformations, and,
    \item[(ii)] the part of the frame that is not composed of moving objects can be considered as stationary background, which typically corresponds to settings with a fixed camera.
\end{itemize}
Such a simplification for motion modelling enables us to combine the motion of an object for multiple time-steps easily and eventually impose time-consistency.


\subsection{Frame Decomposition}
Let $\mathbf{f}_t \in \mathbb{R}^{H\times W \times 3}$ be a video frame at time $t$, with height $H$ and width $W$.
We assume that each frame is composed of $K$ visual entities, which can be considered as objects.
We represent the support of these objects with masks $\mathbf{m}_t^{(k)} \in [0,1]$ for $k=1,\dots,K$.
Each mask is designed to indicate the presence of an object in the scene and is inferred via an autoregressive function as in Equation \eqref{eq:mask0}. 
In other words, we treat the video frames as a source of continual information using this autoregressive function so that the object masks can retain information which is otherwise unobservable at a given time step.

\begin{equation}
    \mathbf{m}_t^{(k)} = F_1(\mathbf{f}_t, \mathbf{m}_{1:t-1}^{(k)}) 
    \label{eq:mask0}
\end{equation}

We let more than one of these $K$ masks have nonzero values for a given pixel index, which would imply occlusion at a given spatial location in the frame.
To handle such cases, we first binarize masks via a soft thresholding mechanism as in Equation \eqref{eq:mask1}, with $\alpha$ and $\beta$ in Equation \eqref{eq:mask1_ad} as hyperparameters.

\begin{align}
    \overline{\mathbf{m}}_t^{(k)} &= \mathcal{S}(\mathbf{m}_t^{(k)}) 
    \label{eq:mask1} \\
    \mathcal{S}(\mathbf{m}) &= 1/(1+\exp^{\left(\alpha*(\mathbf{m}-\beta)\right)})
    \label{eq:mask1_ad}
\end{align}

We then order object masks according to their relative depth by the help of a permutation matrix, $\mathbf{P}_t \in \{0,1\}^{K \times K}$, as given in Equation \eqref{eq:mask2}, which can be learned as a function of masks and the current frame, e.g., $\mathbf{P}_t = F_2(\mathbf{m}_t, \mathbf{f}_t)$.

\begin{equation}
    \{\overline{\overline{\mathbf{m}}}_t^{(k)}\}_{k} = \mathbf{P}_t^T \otimes \{\overline{\mathbf{m}}_t^{(k)}\}_{k} 
    \label{eq:mask2}
\end{equation}

The operator $\otimes$ in Equation \eqref{eq:mask2} denotes matrix multiplication with proper dimension expansion and broadcasting, resulting in an operation of dimensions, $[1\times 1\times K\times K] \otimes [H \times W \times K \times 1] = [H \times W \times K]$.
Permutation allows masks to be learned in a way that is independent of the order of processing, in contrast to iterative algorithms, such as \cite{veerapaneni2020entity}.
As a result, the object representations are invariant to the order or processing, in other words, their assignment of index $k$.

Finally, we decide on the occupancy of each pixel via an occlusion mask denoted by $\Gamma_t \in \{0, 1\}^{H \times W \times K}$, which is a function of ordered object masks as described in Equation \eqref{eq:occ0}.

\begin{align}
    \Gamma_t^{(k)} = \gamma (\{\overline{\overline{\mathbf{m}}}_t^{(k)}\}_{k=1}^K) = \begin{cases}
        \overline{\overline{\mathbf{m}}}_t^{(k)} &\text{if} \; k=1 
        \\
        \max\left(0, \; \overline{\overline{\mathbf{m}}}_t^{(k)} \hspace{-4pt} - \hspace{-2pt} \sum_{l=1}^{k-1} \overline{\overline{\mathbf{m}}}_t^{(l)}\right) \hspace{-4pt} &\text{otherwise}
    \end{cases}
    \label{eq:occ0}
\end{align}

The occlusion mask prioritizes the objects that are closer to the camera when the support of multiple objects try to occupy the same pixel.
In a way, the occlusion mask in our pipeline corresponds to the segmentation mask of a given frame as each pixel is attributed to only one entity, either one of the objects or the background.
However, we would like to emphasize that the occlusion mask in our framework is inferred from object masks, which might contain unobservable information in a given frame.

As a result, the pixelwise product of the frame and the occlusion mask induces the current partial appearance of each object, in other words, the segmentation output: $\mathbf{x}_t^{(k)}$, i.e., $\mathbf{x}_t^{(k)} = \mathbf{f}_t \odot \Gamma_t^{(k)}$.
The rest of the pixels in a frame that are not assigned to any object are then labeled as background, $\mathbf{x}_t^{(K+1)}$ through the binary background mask $\mathbf{m}_t^{(K+1)}$, as given in Equation \eqref{eq:bgnd}, where $\mathbf{1}$ denotes an image of 1's with height $H$ and width $W$.

\begin{equation}
    \mathbf{x}_t^{(K+1)} = \mathbf{f}_t \odot \mathbf{m}_t^{(K+1)} = \mathbf{f}_t \odot \left( \mathbf{1} -  \sum_{k=1}^K \Gamma_t^{(k)} \right)
    \label{eq:bgnd}
\end{equation}

\begin{figure}[t]
	\centering
	\includegraphics[width=0.8\linewidth]{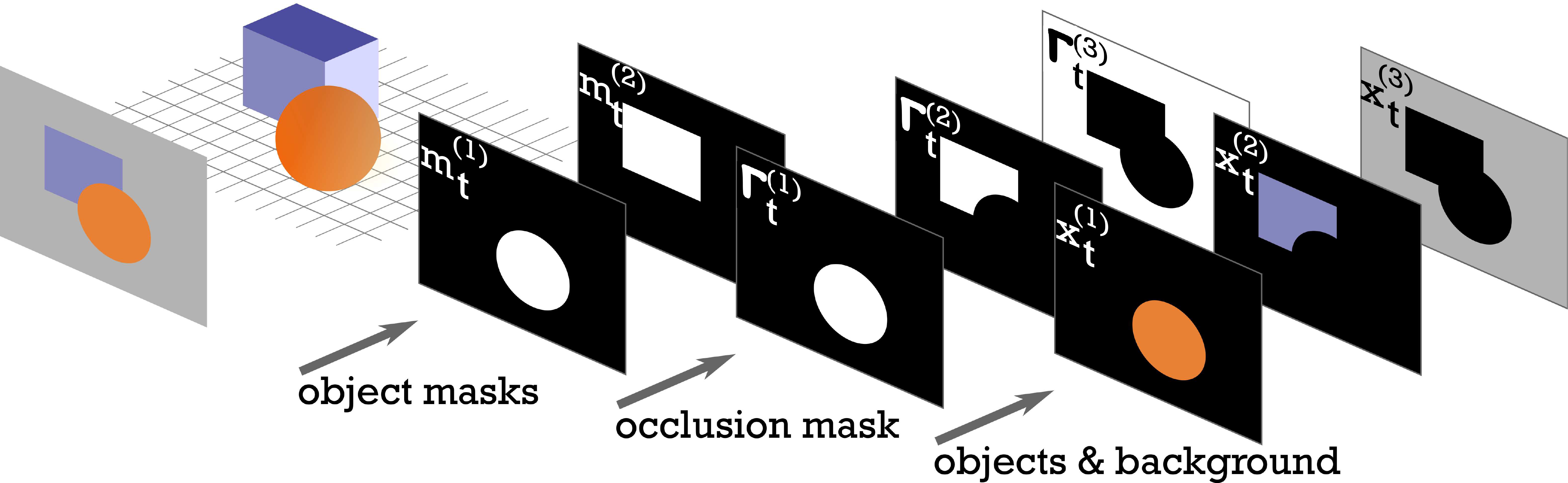}
	\caption{Illustration of proposed frame decomposition.}
	\label{fig:3d}
\end{figure}

Figure \ref{fig:3d} illustrates the notions in our decomposition model for a scene with two objects.
For the illustrated case, the sphere occludes the rectangular prism when the scene is projected onto the 2D plane of the given image.
Note that the proposed object masks aim to cover the full support of each object, when they are projected onto the image plane individually.
However, it is not trivial to infer them when the only source of information is a static image.
On the other hand, a sequence of images where the sphere and the prism in Figure \ref{fig:3d} move in different directions or with different speeds provides richer information for the inference of forenamed object masks.
And once we infer the object masks, all other components of the decomposition module can be computed in a straightforward way.

\subsection{Temporal Transformation Modelling}
For the composition of the frame to be predicted, we need to morph each object consistently with its ongoing motion.
We approximate the temporal evolution of each object individually by planar transformations, such as affine or projective transformation, and denote the corresponding $3\times3$ transformation matrix as $\mathbf{z}^{(k)}_t$.
We predict these object-specific parameters using a nonlinear function as given in Equation \eqref{eq:param0}, whose input includes the current frame, $\mathbf{f}_t$ and previous transformation parameters $\mathbf{z}_{t-1}^{(k)}$, as we believe that motion should be content-dependent and it is usually smooth with respect to time. 

\begin{equation}
    \mathbf{z}_t^{(k)} = F_3 (\mathbf{f}_t, \mathbf{m}_t^{(k)} \hspace{-0.4em}, \mathbf{m}_{t-1}^{(k)}, \mathbf{z}_{t-1}^{(k)}), \quad  \mathbf{z}_{0}^{(k)}:\text{identity} \label{eq:param0}
\end{equation}

Using Spatial Transformers \cite{jaderberg2015spatial}, represented by the operation $\mathcal{T}(\cdot)$, we warp the pixel coordinates for each object with $\mathbf{z}^{(k)}_t$ and use bilinear interpolation to project a mask or object to the next time step.
We use the notation $\mathcal{T}_{\mathbf{z}_{t}^{(k)}} (\cdot)$ to emphasize the parameters of transformation.
More concretely, the warping grid at time $t+1$ for each object can be obtained by Equation \eqref{eq:trans0}, where the pair $(x,y)$ denotes pixel indices.

\begin{equation}
	\begin{bmatrix} x_{t+1}^{(k)} \\ y_{t+1}^{(k)} \\ 1 \end{bmatrix} = \mathbf{z}_t^{(k)} \begin{bmatrix} x_{t}^{(k)} \\ y_{t}^{(k)} \\ 1 \end{bmatrix} 
	\label{eq:trans0}
\end{equation}

By applying bilinear interpolation to each visual input using the transformed pixel locations in Equation \eqref{eq:trans0}, we can predict the appearance of each object as well as its support in the next frame.
Equation \eqref{eq:trans1} describes the bilinear interpolation for object masks, i.e., the interpolation step for the operation $\mathcal{T}_{\mathbf{z}_{t}^{(k)}} (\mathbf{m}_t^{(k)})$.

\begin{equation}
\widehat{\mathbf{m}}_{t+1}^{(k)}(x_{t+1}^{(k)},y_{t+1}^{(k)}) = \sum_{x=1}^{W} \sum_{y=1}^{H} \mathbf{m}_{t}^{(k)}\hspace{-2pt}(x,y) \, \max(0,1-|x_{t}^{(k)}-x|) \, \max(0,1-|y_{t}^{(k)}-y|)
\label{eq:trans1}
\end{equation}


\subsection{Inpainting}
When the objects are transformed using the predicted parameters, the new structure of the scene might reveal previously unseen parts of an object $\mathbf{x}_t^{(k)}$, or of the background $\mathbf{x}_t^{(K+1)}$.
It might also hide some observable parts, which we can expect to see later in the sequence.
Hence, we keep a memory of each object as $\widetilde{\mathbf{x}}_{t}^{(k)}$ via an inpainting module that first transforms the visual clues from previous time step, $\widetilde{\mathbf{x}}_{t-1}^{(k)}$, to time $t$ with the motion parameters predicted at time $t-1$.
This can be considered as registration of object appearance in time, and it is denoted by the operator $\mathcal{T}_{\mathbf{z}_{t-1}^{(k)}} (\widetilde{\mathbf{x}}_{t-1}^{(k)})$ in Equation \eqref{eq:obj}.
The inpainting module then does the update with the current appearance information, $\mathbf{x}_t^{(k)}$ by a simple copy operation from previous object model after time registration.

\begin{equation}
    \widetilde{\mathbf{x}}_{t}^{(k)} = F_4\left(\mathbf{x}_{t}^{(k)}, \mathcal{T}_{\mathbf{z}_{t-1}^{(k)}} (\widetilde{\mathbf{x}}_{t-1}^{(k)})\right), \quad\quad \widetilde{\mathbf{x}}_{0}^{(k)} = \mathbf{x}_{0}^{(k)} 
    \label{eq:obj}
\end{equation}

It is worth noting that such an approach for inpainting results in accumulation of information throughout time and it is a causal operation.
In other words, this so-called inpainting function based on time registered copy operation does not allow previously unseen parts of objects to be "hallucinated"; however, it allows the framework to retain appearance information for each object in case of occlusions, so that the frames after occlusion can be composed based on this accumulated information.

On the other hand, the static background might contain stationary objects, or can be heavily shaded or textured.
In such a case, copying the values of neighboring pixels is more likely to be insufficient. 
Hence, we prefer to inpaint the full appearance of the background via a separate inpainting module, as given in Equation \eqref{eq:bgnd2}.

\begin{equation}
    \widetilde{\mathbf{x}}_{t}^{(K+1)} = F_5\left(\mathbf{x}_{t}^{(K+1)}, \widetilde{\mathbf{x}}_{t-1}^{(K+1)} \right), \quad \widetilde{\mathbf{x}}_{0}^{(K+1)} = \mathbf{x}_{0}^{(K+1)} 
	\label{eq:bgnd2}
\end{equation}


\subsection{Prediction}
Once the current frame, $\mathbf{f}_t$ is decomposed into objects via masks, and object and background models are updated with the inpainting mechanisms as described in the previous part, the task is to project each object to the next time step with the predicted transformation parameters in order to compose the predicted frame, $\widehat{\mathbf{f}}_{t+1}$.
The warping of each inpainted object to time $t+1$ with corresponding parameters is achieved via Equation \eqref{eq:transformer}, where $\widehat{\mathbf{x}}_{t+1}$ refers to inpainted objects after projection to the next time step.

\begin{equation}
    \widehat{\mathbf{x}}_{t+1}^{(k)} = \mathcal{T}_{\mathbf{z}_t^{(k)}} \hspace{-4pt} \left( \widetilde{\mathbf{x}}_{t}^{(k)} \right) 
    \label{eq:transformer}
\end{equation}

In order to infer the occlusion mask at next time step, we also project object masks to the next time step using the same parameters as described in Equation \eqref{eq:occ2}.
The function $\gamma (\cdot)$ is the same as in Equation \eqref{eq:occ0}.

\begin{equation}
    \widehat{\Gamma}_{t+1} = \gamma \hspace{-2pt} \left( \mathbf{P}_t^T \otimes \left\{\mathcal{T}_{\mathbf{z}_{t}^{(k)}} \hspace{-4pt}\left( \overline{\mathbf{m}}_{t}^{(k)}\right)\right\}_{k=1}^K \right)
    \label{eq:occ2} 
\end{equation}

We compose the predicted frame using the predicted background, given in Equation \eqref{eq:bgnd4}, and the transformed objects after the pixelwise product with the occlusion mask predicted for the next step frame, as given in Equation \eqref{eq:compose}.

\begin{align}
    \widehat{\mathbf{x}}_{t+1}^{(K+1)}  &= \widetilde{\mathbf{x}}_{t}^{(K+1)}  \odot \left( \mathbf{1} - \sum_{k=1}^K \widehat{\Gamma}_{t+1}^{(k)} \right)
	\label{eq:bgnd4} \\
	\widehat{\mathbf{f}}_{t+1} &= \bigcup_{k=1}^{K} \left(\widehat{\Gamma}_{t+1}^{(k)} \odot \widehat{\mathbf{x}}_{t+1}^{(k)}\right) \cup \widehat{\mathbf{x}}_{t+1}^{(K+1)} 
	\label{eq:compose} 
\end{align}

As the occlusion mask and the resultant background masks do not overlap by construction, the union operations in Equation \eqref{eq:compose} can be easily implemented as pixelwise summation.

The computational flow for the proposed framework can be found in the Algorithm \ref{algo:framework}.

\section{Learning Framework}
\label{sec:implementation}
Having formulated the next frame prediction problem in an object-centric manner in Section \ref{sec:formulation}, we provide the implementation details in this section, and present our training objective as a combination of prediction and frame decomposition sub-tasks.

\subsection{Implementation}

\begin{figure}[b]
    \centering
    \includegraphics[width=0.8\linewidth]{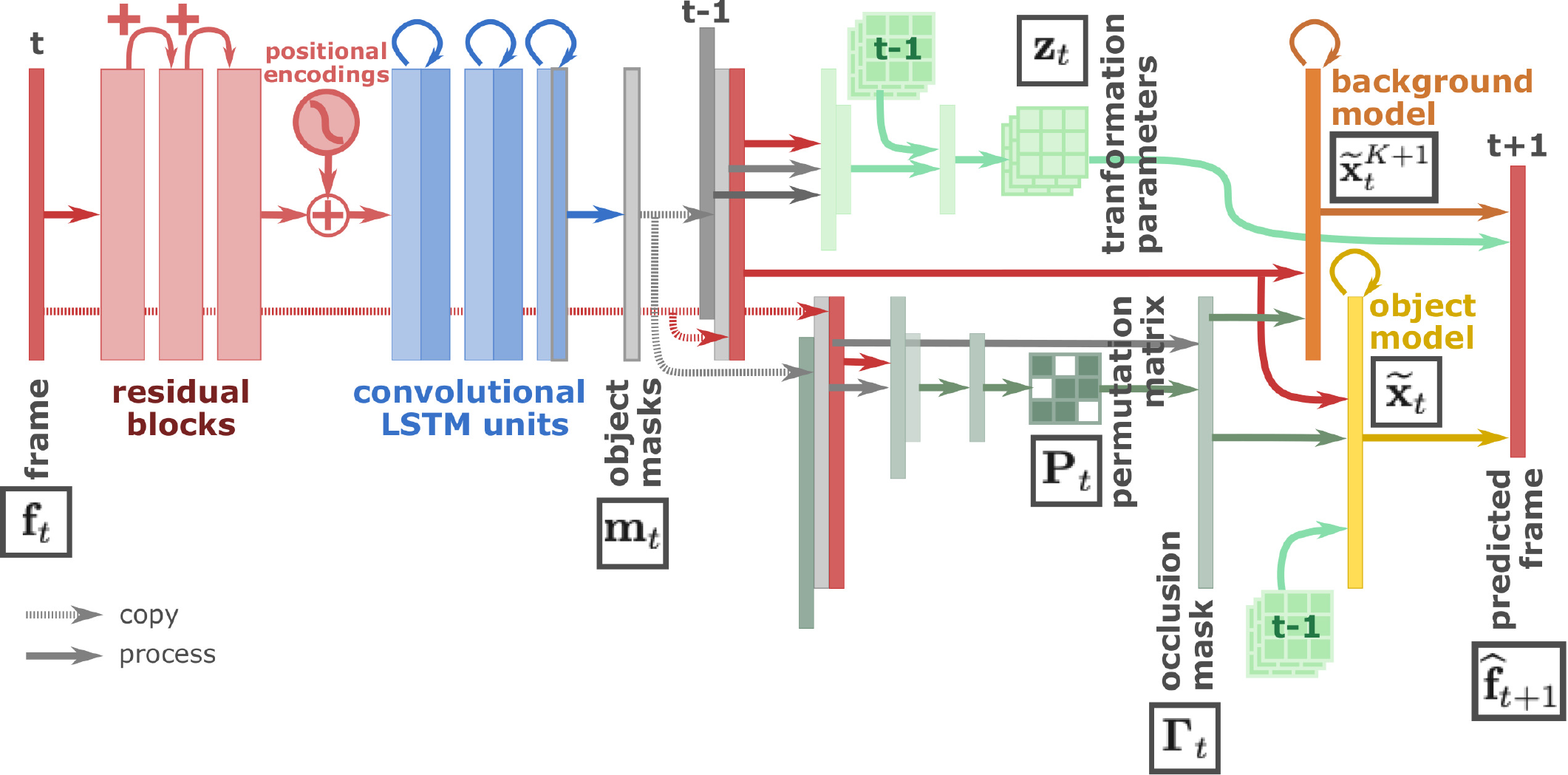}
    \caption{Proposed framework: masks $\mathbf{m}_t^{(k)}$ as output of the recurrent units are used to predict transformation parameters $\mathbf{z}_t^{(k)}$ and relative ordering of objects $\mathbf{P}_t$, which are all then used to compose the predicted frame with occlusion mask $\mathbf{\Gamma}_t$ and inpainting modules for the objects and the background that output $\widetilde{\mathbf{x}}_t$ and $\widetilde{\mathbf{x}}_t^{(K+1)}$, respectively.}
    \label{fig:method}
\end{figure}

For frame decompositon, the autoregressive function $F_1(\cdot)$ in Equation \eqref{eq:mask0} can be implemented in various ways, and we choose to use convolutional LSTM units \cite{xingjian2015convolutional} as they take the spatial structure into account while containing a memory unit.
In order to enrich the input to the LSTM units, we apply residual convolutional layers \cite{he2016deep} to the input frame.
However, as they are translation-invariant by nature, convolutional layers might have difficulties in understanding the global structure in a frame, and recent work \cite{vaswani2017attention} has shown that the positional encodings help neural networks to learn position-aware representations.
Hence, we add sinusoidal positional encodings in 2D as in \cite{wang2020translating} to the embedding of the frame obtained by residual layers, before passing it to the convolutional LSTM units. 

The permutation matrix is approximated as a doubly stochastic matrix (DSM), whose rows and columns sum up to 1, using another nonlinear function with object and occlusion masks as its inputs, i.e., $\mathbf{P}_t = F_2(\mathbf{m}_t, \widehat{\Gamma}_t, \mathbf{f}_t)$.
The function $F_2(\cdot)$ denotes a convolutional neural network (CNN) applied to each object separately, which is followed by fully connected layers (MLP), with the uppermost layer having $K$ hidden units.
The outputs of the MLP for each object is then used to construct a $K \times K$ matrix, which is validated to be a DSM via the Gumbel-Sinkhorn algorithm \cite{mena2018learning}.

For the prediction of transformation parameters, we use another CNN to process visual input of $F_3(\cdot)$ in Equation \eqref{eq:param0}, and concatenate its output to the parameters from previous time step as input to a second MLP.
The convolutional kernels and weights in MLPs for permutation learning and parameter prediction are shared across all objects, which makes the model scalable with increasing number of objects, both in terms of computation and memory, in contrast to existing iterative algorithms \cite{zablotskaia2020unsupervised, veerapaneni2020entity}.

For object inpainting, the function $F_4(\cdot)$ in Equation \eqref{eq:obj} is a selection operator based on a if-statement to copy values of pixels in occluded regions from previous frames, whereas $F_5(\cdot)$ in Equation \eqref{eq:bgnd2} is implemented as fully convolutional networks that use the same kernels for R,G,B channels.

The whole end-to-end differentiable framework is illustrated in Figure \ref{fig:method}.

\begin{algorithm}[t]
\DontPrintSemicolon
\KwInput{Video sequence: $\{\mathbf{f}_t\}_{t=1}^T$, number of moving objects to be inferred: $K$}
\KwOutput{Predicted sequence: $\{\widehat{\mathbf{f}}_t\}_{t=1}^{T-1}$; object masks $\{\mathbf{m}_t^{(k)}\}_{t=1}^{T-1}$ and inpainted objects $\{\widetilde{\mathbf{x}}_t^{(k)}\}_{t=1}^{T-1}$ for $k=1,\dots,K$; inpainted background $\{\widetilde{\mathbf{x}}_t^{(K+1)}\}_{t=1}^{T-1}$}
\KwInit{$\mathbf{z}_0^{(k)}=\text{identity}$}

\vspace{0.25em}
\For{$t=1,\dots,T-1$}
{
    \tcc{Decompose the frame}
    $\mathbf{m}_t^{(k)} \leftarrow F_1(\mathbf{f}_t, \mathbf{m}_{1:t-1}^{(k)})$ 
    \tcp*{Equation \eqref{eq:mask0}}
    
    \vspace{0.25em}
    \tcc{Determine the permutation matrix}
    \If{ $t=1$ } 
        {$\widehat{\Gamma}_t = \gamma ( S (\{\mathbf{m}_t^{(k)}\}_{k=1}^K) ) $
        \tcp*{Equations \eqref{eq:mask1},\eqref{eq:mask1_ad},\eqref{eq:occ0}}}
    $\mathbf{P}_t \leftarrow F_2(\mathbf{m}_t, \widehat{\Gamma}_t, \mathbf{f}_t)$
    
    \vspace{0.25em}
    \tcc{Binarize object masks, order them and generate the occlusion mask}
    ${\Gamma}_t \leftarrow \gamma \left( \mathbf{P}_t^T \otimes \{S(\mathbf{m}_t^{(k)}\}_{k} ) \right)$
    \tcp*{Equations \eqref{eq:mask1}-\eqref{eq:occ0}}
    
    \vspace{0.25em}
    \tcc{Infer the objects and the background}
    $\mathbf{x}_t^{(k)} \leftarrow \mathbf{f}_t \odot \Gamma_t^{(k)} \quad k=1,\dots,K$ \\
    $\mathbf{x}_t^{(K+1)} \leftarrow \mathbf{f}_t \odot \mathbf{m}_t^{(K+1)}$
    \tcp*{Equation \eqref{eq:bgnd}}
    
    \vspace{0.25em}
    \tcc{Inpaint the objects and the background}
    \If{ $t=1$ } 
        {$\widetilde{\mathbf{x}}_{t}^{(k)} \leftarrow \mathbf{x}_t^{(k)} \quad k=1,\dots,K$ \\
        $\widetilde{\mathbf{x}}_{t}^{(K+1)} \leftarrow \mathbf{x}_t^{(K+1)}$}
    \Else
        {$\widetilde{\mathbf{x}}_{t}^{(k)} \leftarrow F_4\left(\mathbf{x}_{t}^{(k)}, \mathcal{T}_{\mathbf{z}_{t-1}^{(k)}} (\widetilde{\mathbf{x}}_{t-1}^{(k)})\right) \quad k=1,\dots,K$
        \tcp*{Equation \eqref{eq:obj}}
        $\widetilde{\mathbf{x}}_{t}^{(K+1)} \leftarrow F_5\left(\mathbf{x}_{t}^{(K+1)}, \widetilde{\mathbf{x}}_{t-1}^{(K+1)} \right)$
        \tcp*{Equation \eqref{eq:bgnd2}}}
        
    \vspace{0.25em}
    \tcc{Predict parameters for the geometric transformation}
    \If{ $t=1$ }
    {$\mathbf{z}_t^{(k)} \leftarrow F_3 (\mathbf{f}_t, \mathbf{m}_t^{(k)} \hspace{-0.4em}, \mathbf{m}_{t}^{(k)}, \mathbf{z}_{t-1}^{(k)})$}
    \Else
    {$\mathbf{z}_t^{(k)} \leftarrow F_3 (\mathbf{f}_t, \mathbf{m}_t^{(k)} \hspace{-0.4em}, \mathbf{m}_{t-1}^{(k)}, \mathbf{z}_{t-1}^{(k)})$ 
    \tcp*{Equation \eqref{eq:param0}}}
    
    \vspace{0.25em}
    \tcc{Transform inpainted objects}
    $\widehat{\mathbf{x}}_{t+1}^{(k)} \leftarrow \mathcal{T}_{\mathbf{z}_t^{(k)}} \hspace{-4pt} \left( \widetilde{\mathbf{x}}_{t}^{(k)} \right) $
    \tcp*{Equation \eqref{eq:transformer}}
    
    \vspace{0.25em}
    \tcc{Transform and order binarized masks, predict the occlusion mask for the next frame \hspace{-2em}}
    $\widehat{\Gamma}_{t+1} \leftarrow \gamma \hspace{-2pt} \left( \mathbf{P}_t^T \otimes \left\{\mathcal{T}_{\mathbf{z}_{t}^{(k)}} \hspace{-4pt}\left( \overline{\mathbf{m}}_{t}^{(k)}\right)\right\}_{k=1}^K \right)$
    \tcp*{Equation \eqref{eq:occ2}}
    
    \vspace{0.25em}
    \tcc{Compose the predicted frame}
    $\widehat{\mathbf{f}}_{t+1} \leftarrow \bigcup_{k=1}^{K} \left(\widehat{\Gamma}_{t+1}^{(k)} \odot \widehat{\mathbf{x}}_{t+1}^{(k)}\right) \cup \left( \widetilde{\mathbf{x}}_{t}^{(K+1)} \odot \left( \mathbf{1} - \sum_{k=1}^K \widehat{\Gamma}_{t+1}^{(k)} \right) \right)$
	\tcp*{Equations \eqref{eq:bgnd4},\eqref{eq:compose}}
    }
    
\caption{Algorithm for object-centric next frame prediction in videos}
\label{algo:framework}
\end{algorithm}

\subsection{Loss}
The joint learning of all modules is achieved via the minimization of the reconstruction error between the predicted frame, $\widehat{\mathbf{f}}_{t+1}$, and the ground truth next frame, $\mathbf{f}_{t+1}$, at each time step $t$, which is later averaged over the whole sequence of length $T$, as in Equation \eqref{eq:recon}, where $\|.\|_1$ denotes $L1$ norm.

\begin{equation}
    \mathcal{L}_{p} = \frac{1}{T-1}\sum_{t=1}^{T-1} \|\mathbf{f}_{t+1} - \widehat{\mathbf{f}}_{t+1} \|_1 \label{eq:recon}
\end{equation}

In order to prevent the trivial solution where the network ends up representing the whole frame as a single object, we introduce auxiliary loss terms; namely, $\mathcal{L}_{m1}$ and $\mathcal{L}_{m2}$, which promote sparsity and compactness of the object masks, respectively, and $\mathcal{L}_{c}$ which verifies the consistency of predicted spatial transformations in time.
We minimize the weighted sum given in Equation \eqref{eq:loss}, where $\{\lambda_i\}_{i=1}^5$ are hyperparameters, $\theta$ represents all learnable parameters and $\|.\|_2$ denotes $L2$ norm.
We use the rmsprop algorithm in \cite{kingma2014adam} as optimizer to minimize the total loss $\mathcal{L}$ via batch gradient descent.

\begin{equation}
    \mathcal{L} = \lambda_1 \mathcal{L}_{p} + \lambda_2 \mathcal{L}_{m1} + \lambda_3 \mathcal{L}_{m2} + \lambda_4 \mathcal{L}_{c} + \lambda_5 \|\theta\|_2^2
    \label{eq:loss}
\end{equation}

The details of auxiliary loss terms are as follows:

\vspace{1em}
\noindent \textbf{Mask Sparsity Loss:} For spatially sparse representations, we penalize the average $L_1$ norm of the object masks as in Equation \eqref{eq:sparsity}, where $(x,y)$ represents pixel coordinates.

\begin{equation}
    \mathcal{L}_{m1} = \frac{1}{HWK} \sum_k \sum_{x,y} \left( \|\mathbf{m}^{(k)}(x,y)\|_1 \right)
    \label{eq:sparsity} 
\end{equation}

\noindent \textbf{Mask Concentration Loss:} The sparsity constraint for object masks might be insufficient alone as each mask is also expected to be spatially connected, in general.
In order to promote compactness for object masks, we further use a geometric concentration loss as given in Equation \eqref{eq:concent}, which was originally proposed for unsupervised segmentation problem \cite{hung2019scops}.
The tuple $(x,y)$ stands for pixel coordinates as before, and $(c_x^{(k)}, c_y^{(k)})$ denotes the center of mass for each mask and it is computed as described in Equation \eqref{eq:concent}.

\begin{equation}
    \mathcal{L}_{m2} = \sum_k \sum_{x,y} \left(\hspace{-2pt}\| (x,y) \hspace{-2pt} - \hspace{-2pt} (c_x^{(k)}, c_y^{(k)}) \|_2^2 \frac{\mathbf{m}^{(k)}(x,y)}{\sum_{x,y} \mathbf{m}^{(k)}(x,y)} \right) 
    \label{eq:concent} 
\end{equation}

\begin{equation}
    c_x^{(k)} = \sum_{x,y} x \frac{\mathbf{m}^{(k)}(x,y)}{\sum_{x,y} \mathbf{m}^{(k)}(x,y)}, \quad
    c_y^{(k)} = \sum_{x,y} y \frac{\mathbf{m}^{(k)}(x,y)}{\sum_{x,y} \mathbf{m}^{(k)}(x,y)}
    \label{eq:concent2} 
\end{equation}

\noindent \textbf{Consistency Loss:} The use of geometric transformations enables us to combine estimated motion between any given pair of time instances via matrix multiplication of parameters. 
One straightforward implementation to exploit this fact would be the minimization of reconstruction error between any $(t_1', t_2')$ time instant pairs, where $t_1', t_2' \in \{0, T\}$ with $|t_1' - t_2'| > 1$, where $|\cdot|$ denotes absolute value.
In order to keep the computational cost low, and imitate a similar reconstruction penalty based on consistency, we sample two time instants $t_1 \sim U[0, T/2]$ and $t_2 \sim U(T/2, T]$ at each training step, where $U[\cdot, \cdot]$ resembles discrete uniform distribution.
By reconstructing the frame at $t_2$ using the masks and objects inferred for time $t_1$, and also doing it backwards, i.e., reconstructing the frame at $t_1$ using the masks and objects inferred for time $t_2$, we can cover the reconstruction penalty emerging from all aforementioned $(t_1', t_2')$ time instant pairs as long as we realize enough training steps.

For the first component of the loss, we predict $\mathbf{f}_{t_2}$ using $\mathbf{m}_{t_1}, \{\mathbf{z}_t\}_{t=t_1}^{t_2-1}, \mathbf{P}_{t_2 - 1}$ and $\{\widetilde{\mathbf{x}}_{t_1}^{(k)} \}_{k=1}^{K+1}$.
We first extrapolate motion as in Equation \eqref{eq:cons0} and use the notation $\circ$ to represent the combination of parameters to project the masks and objects from $t_1$ to $t_2$.

\begin{equation}
    \mathbf{z}_{\circ}^{(k)} = \mathbf{z}_{t_2-1}^{(k)} \, \mathbf{z}_{t_2-2}^{(k)} \dots \mathbf{z}_{t_1}^{(k)} \; \forall k
    \label{eq:cons0}
\end{equation}

We use the parameters $\mathbf{z}_{\circ}^{(k)}$ to transform binarized object masks from time $t_1$ to $t_2$, 
Upon projecting, we order the masks according to their relative depth using the permutation matrix $\mathbf{P}_{t_2 - 1}$ as in Equation \eqref{eq:cons1}, resulting in $\{\widehat{\mathbf{m}}_{\circ}^{(k)}\}$.
The reason behind the choice of permutation matrix at time $t_2 -1$ rather than $t_1$ is that the relative ordering of objects can change in time and the ordering at the previous time step is more relevant than the ordering multiple time steps back.

\begin{equation}
    \{\widehat{\mathbf{m}}_{\circ}^{(k)}\}_k = \mathbf{P}_{t_2-1}^T \otimes \left\{\mathcal{T}_{\mathbf{z}_{\circ}^{k}} \hspace{-2pt} \left(S(\mathbf{m}_{t_1}^{(k)})\right)\right\}_k \label{eq:cons1}
\end{equation}

Once the ordering is realized, we predict the occlusion mask for time $t_2$ as given in Equation \eqref{eq:cons2}, where $\gamma(\cdot)$ is the same function described in Equation \eqref{eq:occ0}.

\begin{equation}
    \Gamma_{\circ}^{(k)} = \gamma \hspace{-2pt} \left( \{\widehat{\mathbf{m}}_{\circ}^{(k)}\}_{k=1}^K \right) 
    \label{eq:cons2}
\end{equation}

Using our composition principles, we generate $\widehat{\mathbf{f}}_{\circ}$, the prediction of the frame at time $t_2$,  as described in Equation \eqref{eq:cons3}.

\begin{equation}
    \widehat{\mathbf{f}}_{\circ} \hspace{-2pt} = \hspace{-2pt} \left(\bigcup_{k=1}^{K} \Gamma_{\circ}^{(k)} \hspace{-2pt} \odot \hspace{-2pt} \mathcal{T}_{\overline{\mathbf{z}}_{\circ}^{k}} \hspace{-2pt} \left(\widetilde{\mathbf{x}}_{t_1}^{(k)}\right) \hspace{-4pt} \right) \hspace{-2pt} \cup \hspace{-2pt} \left( \hspace{-4pt} \left( \hspace{-2pt} \mathbf{1} \hspace{-2pt} - \hspace{-4pt} \sum_{k=1}^K \Gamma_{\circ}^{(k)} \hspace{-2pt} \right) \hspace{-2pt} \odot \hspace{-2pt} \widetilde{\mathbf{x}}_{t_1}^{(K+1)} \hspace{-2pt} \right) \label{eq:cons3}
\end{equation}

Finally, in order to have symmetry in time and provide more information for learning masks and permutation, we also use the inverse of the combined parameters, $\mathbf{z}_{\bullet}^{(k)} = {\mathbf{z}_{\circ}^{(k)}}^{-1}$ for the backward prediction in time.
We use the notation $\bullet$ to represent the combination of parameters to project the masks and objects from $t_2$ to $t_1$.
Using  $\mathbf{z}_{\bullet}^{(k)}, \mathbf{m}_{t_2}, \mathbf{P}_{t_1 + 1}$ and $\{\widetilde{\mathbf{x}}_{t_2}^{(k)} \}_{k=1}^{K+1}$ this time, we follow the steps in Equation \eqref{eq:cons0}-\eqref{eq:cons3} to reverse predict the frame at time $t_1$ as $\widehat{\mathbf{f}}_{\bullet}$.
We finally define the cyclic time consistency loss as the average of the prediction errors from $t_1$ to $t_2$ and from $t_2$ to $t_1$, as given in Equation  \eqref{eq:cons_loss}. 

\begin{equation}
\mathcal{L}_{c} = \frac{\| \mathbf{f}_{t_2} - \widehat{\mathbf{f}}_{\circ} \|_1 + \| \mathbf{f}_{t_1} - \widehat{\mathbf{f}}_{\bullet} \|_1}{2}
\label{eq:cons_loss}
\end{equation}

We believe that the cyclic consistency loss does not only assists the transformation parameters to evolve from identity to the ground truth values, but it also aids the memory units for object masks to retain the partially observable information in case of occlusions.

\section{Experiments}
\label{sec:experiments}
We evaluate the performance of our framework on a simulated dataset.
We generate sequences with $T=11$ frames of size $64\times64$, where $K_d$ objects start moving from randomly chosen quadrants of the frame towards the opposite quadrant with a random speed, which induces occlusion later in each sequence.
The initial position in the quadrant and the order of objects when they occlude each other are also chosen randomly.
Objects are randomly selected from a \verb|{circle, square, triangle}| shape set with $5^3-1$ different color possibilities.

\begin{figure}[h]
	\centering
	\includegraphics[height=4cm]{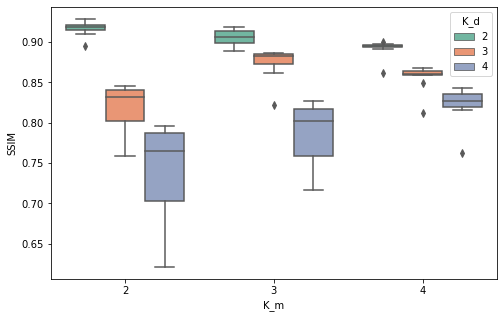} 
	\hspace{1cm}
	\includegraphics[height=4cm]{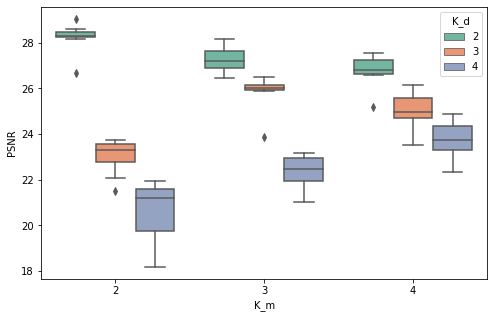} 
	\caption{The prediction ability of the model presented as SSIM and PSNR metrics for different numbers of objects in the model ($K_m$) and in the test dataset ($K_m$). The variation in each boxplot, i.e., a specific trained model tested on a specific dataset, depicts the variation in prediction qulity for each time step at a given sequence.}
    \label{fig:quan}
\end{figure}

We generate three datasets with $K_d \in \{2,3,4\}$ objects present in a frame.
For each case, the training set consists of 20K sequences whereas the validation and the test sets each have 2K sequences.
We train three models with $K_m \in \{2,3,4\}$ object representations, and evaluate each model with the three datasets.
In other words, we evaluate the performance of a model when the number of objects in the model and the number of objects present in the dataset match, i.e., $K_m = K_d$, and respectively mismatch, $K_m \neq K_d$.
In Figure \ref{fig:quan}, we report the prediction quality for each case on the test set in terms of peak-to-noise signal ratio (PSNR) and structural similarity (SSIM), which are averaged over the entire test set for each time step.
We present the distribution of the quality metric throughout the sequence, without discarding any frame, although it is a common practice to discard few initial frames as warm-up.
In particular, we note that our model is not exposed to "\textit{moving objects}" at the very first frame. 
It can be clearly observed that the model is capable of generating the predicted frame decently as long as $K_d \leq K_m$.
The deterioration in performance when $K_d > K_m$ is expected as the model has to predict the same motion for multiple objects and ordering gets complicated.
The performance slightly degrades with larger number of objects when $K_d = K_m$, due to the higher complexity arising from heavy occlusions.
However, we clearly see that the model generalizes well for $K_d < K_m$, which indicates that $K_m$ is not critically required a priori and a model trained for a large enough $K_m$ can be used for datsasets with different numbers of objects.

\begin{figure}[b!]
	\centering
	\includegraphics[width=0.6\linewidth]{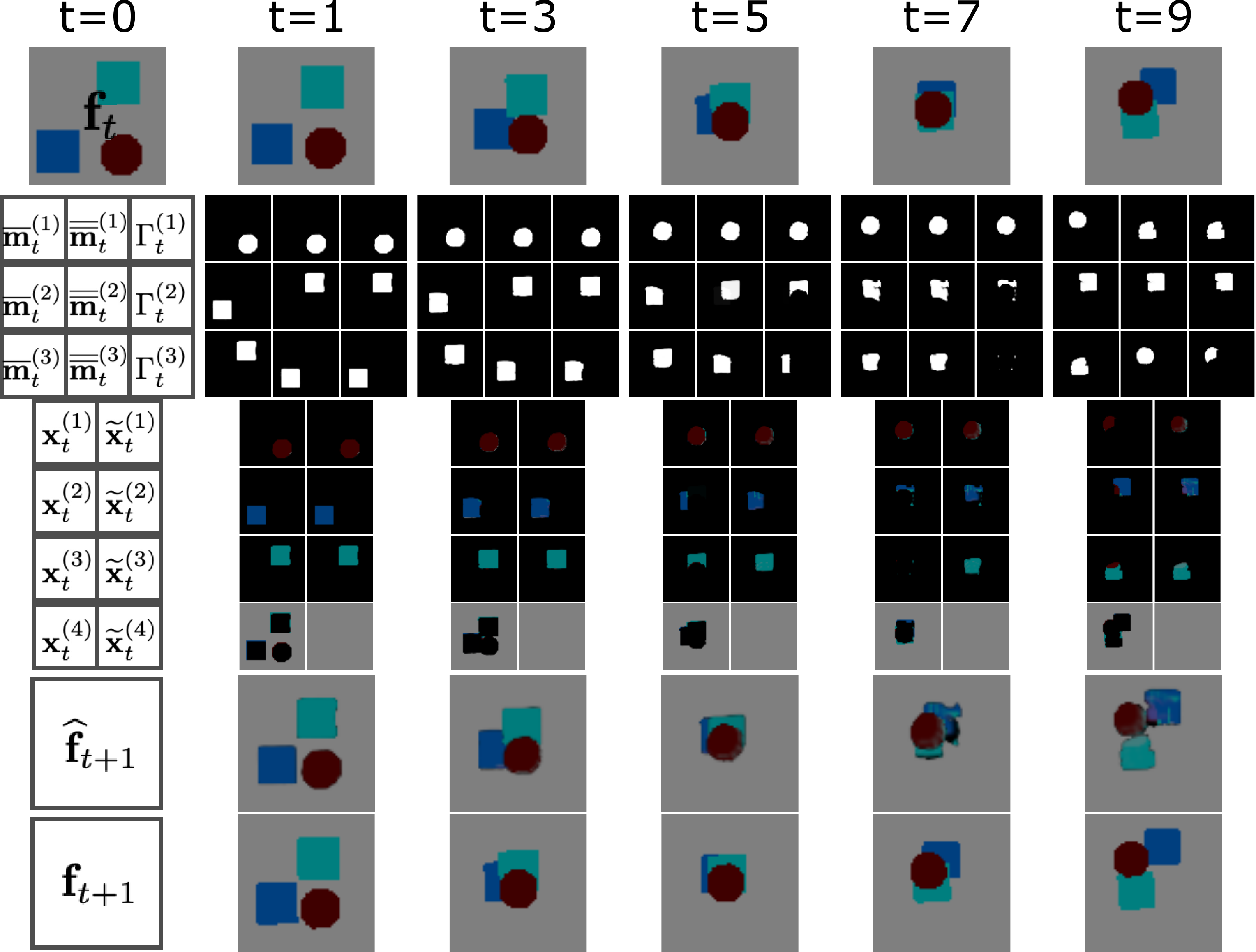}
	\caption{Illustrative output for a test sequence with object masks after binarization ($\overline{\mathbf{m}}_t^{(k)}$), and ordering ($\overline{\overline{\mathbf{m}}}_t^{(k)}$); occlusion mask ($\Gamma_t^{(k)}$), objects and background before and after inpainting ($\{\mathbf{x}_t^{(k)}\}_{k=1}^{K+1}$ and $\{\widetilde{\mathbf{x}}_t^{(k)}\}_{k=1}^{K+1}$ respectively), as well as the ground-truth and predicted frames (${\mathbf{f}}_{t+1}$ and $\widehat{\mathbf{f}}_{t+1}$), for a sequence and trained model with $K_d = K_m = 3$.}
    \label{fig:res1}
\end{figure}

\begin{figure}[h]
	\begin{subfigure}{.33\textwidth}
	    \centering
	    \includegraphics[width=\linewidth]{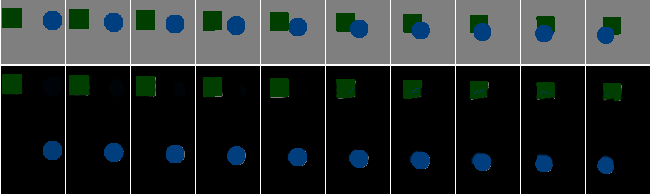}
	    \caption{$K_m = 2, \; K_d = 2$}
	\end{subfigure}
	\begin{subfigure}{.33\textwidth}
	    \centering
	    \includegraphics[width=\linewidth]{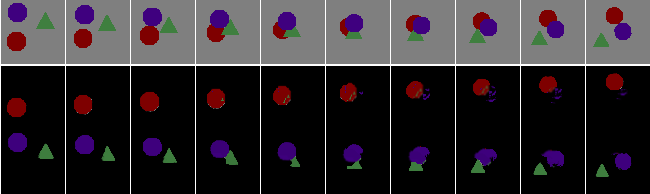}
	\caption{$K_m = 2, \; K_d = 3$}
	\end{subfigure}
	\begin{subfigure}{.33\textwidth}
	    \centering
	    \includegraphics[width=\linewidth]{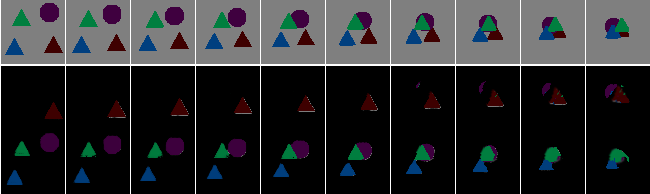}
	\caption{$K_m = 2, \; K_d = 4$}
	\end{subfigure} \\\begin{subfigure}{.33\textwidth}
	    \centering
	    \includegraphics[width=\linewidth]{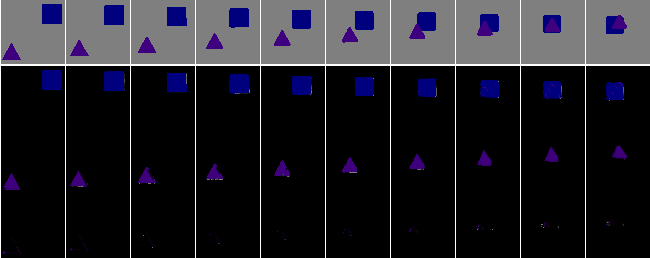}
	    \caption{$K_m = 3, \; K_d = 2$}
	\end{subfigure}
	\begin{subfigure}{.33\textwidth}
	    \centering
	    \includegraphics[width=\linewidth]{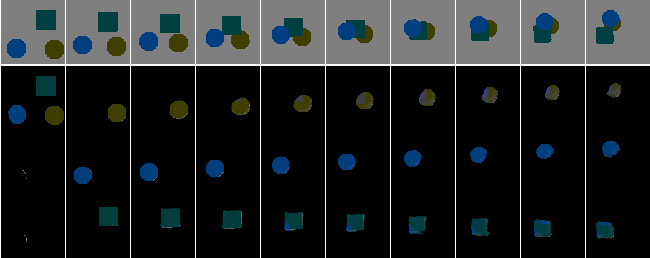}
	\caption{$K_m = 3, \; K_d = 3$}
	\end{subfigure}
	\begin{subfigure}{.33\textwidth}
	    \centering
	    \includegraphics[width=\linewidth]{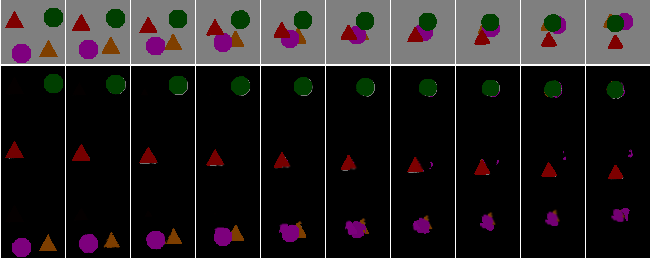}
	\caption{$K_m = 3, \; K_d = 4$}
	\end{subfigure} \\
	\begin{subfigure}{.33\textwidth}
	    \centering
	    \includegraphics[width=\linewidth]{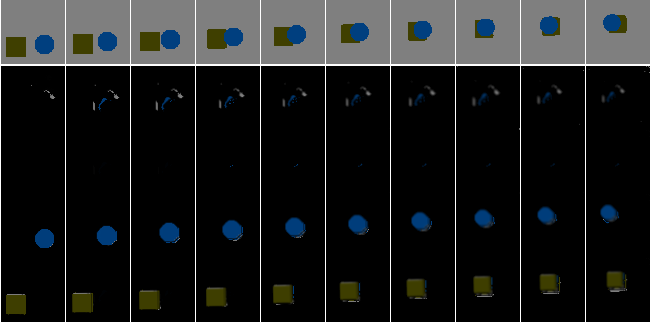}
	    \caption{$K_m = 4, \; K_d = 2$}
	\end{subfigure}
	\begin{subfigure}{.33\textwidth}
	    \centering
	    \includegraphics[width=\linewidth]{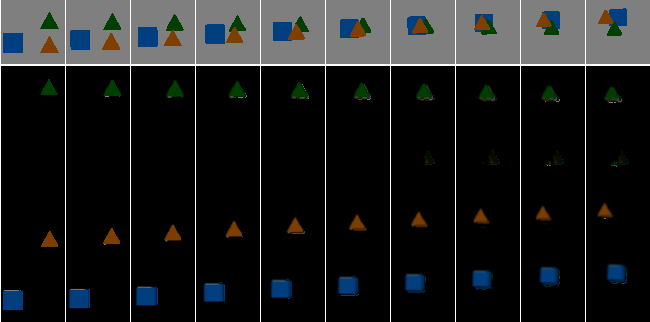}
	\caption{$K_m = 4, \; K_d = 3$}
	\end{subfigure}
	\begin{subfigure}{.33\textwidth}
	    \centering
	    \includegraphics[width=\linewidth]{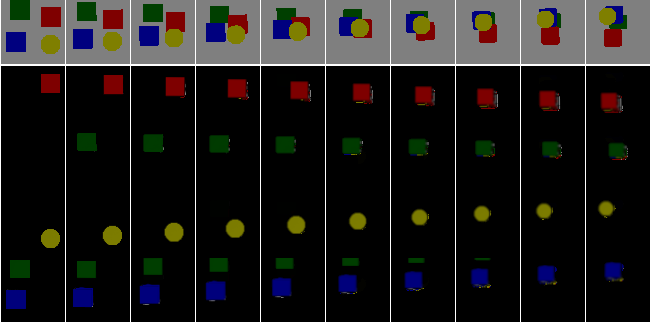}
	\caption{$K_m = 4, \; K_d = 4$}
	\end{subfigure} \\
	\caption{Test sequences $\{\mathbf{f}_t\}_{t=1}^{10}$ with inferred objects $\{\widetilde{\mathbf{x}}_t^{(k)}\}_{t=1}^{10}$ for different $K_m$ and $K_d$. Each row of images correspond to a different $K_m$, a single model, whereas columns represent the test dataset with different number of objects, $K_d$.}
    \label{fig:res2}
\end{figure}

As our claim is not to predict the next frame the best way, but do it in a compositional manner by handling occlusions, and actually use the prediction task as supervision for object discovery, we also provide qualitative results in Figure \ref{fig:res1} and \ref{fig:res2}.
In Figure \ref{fig:res1}, we can observe that the model can infer object shapes as masks, even in case of heavy occlusions, e.g., at time $t=5$ and $t=7$. 
The correct relative ordering of objects is achieved by switching the second and the third object masks for the construction of the occlusion mask.
The partial object appearance obtained as a product of the current frame and occlusion mask is reasonably inpainted by our simple yet effective inpainting module.
For example, even though the third object with square shape and turquoise color is almost fully occluded at time $t=7$, our model can keep a memory of its appearance as illustrated by $\widehat{\mathbf{x}}_7^{(3)}$.
Note that our model does not only recover object appearance after occlusion (e.g., between time $t=5$ to $t=9$ for this particular example), it also correctly positions the objects for the predicted frame by correctly estimating their motion.
Figure \ref{fig:res2} exemplifies more sequences where we demonstrate the objects that our model infers, i.e., $\{\widetilde{\mathbf{x}}_t^{(k)}\}_{k=1}^{K_m}$ for different $K_m$ and $K_d$, without any supervision or pre-trained networks.  
Given that all the sub-problems we are trying to tackle in this work, such as segmentation, inpainting or prediction, are already well-established research problems with their own challenges, even in the presence of annotated datasets, these preliminary results indicate the efficacy of our formulation where each sub-module is designed to aid the others for learning everything from data in a fully unsupervised fashion.


\begin{figure}[h]
	\centering
	\begin{subfigure}{.22\textwidth}
		\centering
		\includegraphics[height=3.2cm]{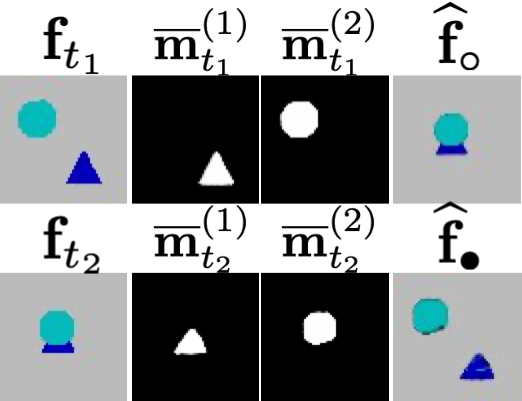}
		\caption{$K_m = K_d = 2$}
	\end{subfigure}
	\hfill
	\begin{subfigure}{.28\textwidth}
		\centering
		\includegraphics[height=3.2cm]{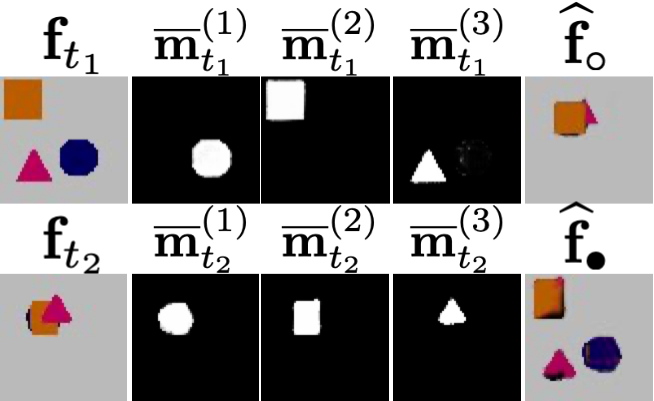}
		\caption{$K_m = K_d = 3$}
	\end{subfigure}
	\hfill
	\begin{subfigure}{.38\textwidth}
		\centering
		\includegraphics[height=3.2cm]{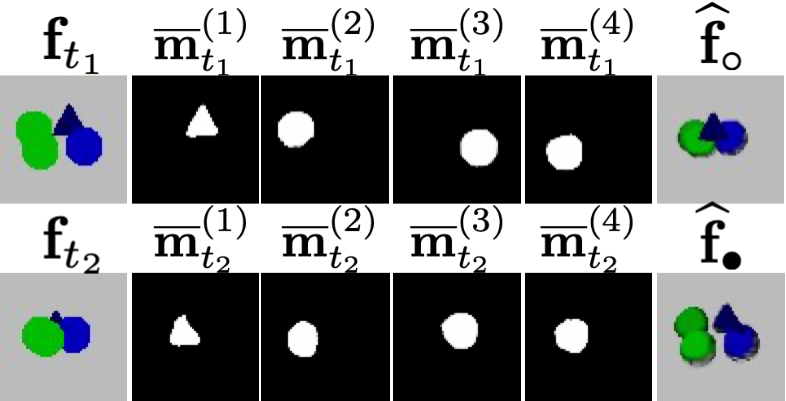}
		\caption{$K_m = K_d = 4$}
	\end{subfigure}
	\caption{Probing on cyclic consistency loss: the top row illustrates the reconstruction from time $t_1$ to $t_2$, where the first image in each block/row represents $\mathbf{f}_{t_1}$, followed by masks inferred by the model, i.e., $\{\overline{\mathbf{m}}_{t_1}^{(k)}\}_{k=1}^{K_m}$ and the prediction $\widehat{\mathbf{f}}_\circ$. The bottom row, on the other hand, illustrates the backward reconstruction time $t_2$ to $t_1$, with $\mathbf{f}_{t_2}$,  $\{\overline{\mathbf{m}}_{t_2}^{(k)}\}_{k=1}^{K_m}$ and $\widehat{\mathbf{f}}_\bullet$. Note that the consistency loss penalizes both reconstructions $(\widehat{\mathbf{f}}_\circ - \mathbf{f}_{t_2})$ and $(\widehat{\mathbf{f}}_\bullet  - \mathbf{f}_{t_1})$.}
	\label{fig:cons}
\end{figure}

In addition, in order to emphasize the importance of cyclic time consistency loss, we select a validation sequence for the three models with $K_m = K_d $, and briefly illustrate in Figure \ref{fig:cons} how this loss term promotes accurate prediction of transformation parameters for multiple time steps while helping the network learn  plausible object masks despite of occlusions.
The first row for each model depicts the prediction of the frame $\mathbf{f}_{t_2}$ as $\widehat{\mathbf{f}}_\circ$ using $\mathbf{f}_{t_1}$ and the object masks inferred at time $t_1$, i.e., $\{\mathbf{m}_{t_1}^{(k)}\}_{k=1}^{K_m}$.
The bottom row for each model visualizes the prediction of the frame $\mathbf{f}_{t_1}$ as $\widehat{\mathbf{f}}_\bullet$ using $\mathbf{f}_{t_2}$ and the object masks inferred at time $t_2$, i.e., $\{\mathbf{m}_{t_2}^{(k)}\}_{k=1}^{K_m}$, which is slightly more interesting as the time step $t_2$ usually corresponds to frames with heavy occlusion in our dataset.
In order to predict the frame $\mathbf{f}_{t_1}$ using the masks and objects inferred at time $t_2$, the model has to understand the structure and remember the shapes of objects reasonably well.
The samples of models $K_m = K_d = 3$ and $K_m = K_d = 4$ also point out that the sub-optimal quantitative performance in Figure \ref{fig:quan} can be partially attributed to a problem with ordering objects.
Such an outcome with clear structure and object shapes with inaccurate depth ordering can be preferred over a blurry outcome of a non object-centric prediction in many applications.

\section{Conclusion}
\label{sec:conclusion}
In this work, we have proposed a novel end-to-end differentiable compositional architecture for video frame representation with explicit occlusion handling.
Using next frame as supervisory signal and employing efficient auxiliary loss terms, we have shown how our framework can explore and represent objects, even when they are heavily occluded.
The experimental results indicate that the number of objects is not required to be known a priori as long as a model with large enough $K$, i.e., number of moving objects, can be trained for a given dataset.
The proposed framework is very generic and does not require any annotations or pre-trained models.
Although our work is currently based on simplifying assumptions, which nevertheless hold in many scenarios with rigid and non-transparent objects moving without in-plane rotations, the framework is very flexible for future extensions.

\clearpage
\label{sec:ref}
\bibliographystyle{IEEEbib}
\bibliography{refs}

\end{document}